# Human-AI Collaboration Increases Efficiency in Regulatory Writing

**Authors**

**Umut Eser[1]\***, Yael Gozin[2], L. Jay Stallons[2], Ari Caroline[1], Martin Preusse[1], Brandon Rice[1], Scott Wright[2], Andrew Robertson[2]

[1] Weave Platform, USA

[2] Takeda Pharmaceuticals, USA

---

**Corresponding Author**

\* Umut Eser

Email: **umut@weave.bio**

---

**Author Emails**

**Weave Platform**:
ari@weave.bio | martin@weave.bio | brandon@weave.bio

**Takeda Pharmaceuticals**:
yael.gozin@takeda.com | lindsey-jay.stallons@takeda.com | scott.wright@takeda.com | andrew.robertson@takeda.com

# Abstract

**Background:** Investigational New Drug (IND) application preparation is a time-intensive, expertise-dependent process that can slow progress and ultimately delay the clinical development timelines.

**Objective:** To evaluate whether a large language model (LLM)-based platform designed for drafting of IND applications can reduce first-draft composition time while maintaining document quality in regulatory submissions.

**Methods: Time efficiency.** Times for IND nonclinical written summaries (eCTD modules 2.6.2, 2.6.4, 2.6.6) generated by AutoIND were directly recorded for all tasks resulted in generating the first drafts. For comparison, drafting times for IND nonclinical written summaries manually generated and cleared by the U.S. Food and Drug Administration's (FDA) during the last 4 years were estimated based on the experience of regulatory writers (≥ 6 years' experience) at a multinational biotechnology company (Takeda Biopharmaceuticals). Manual times were used only as industry-standard benchmarks to contextualize efficiency, while AutoIND times reflect actual measured drafting of complete first drafts.

**Quality assessment.** An experienced regulatory writing assessor evaluated AutoIND-generated drafts using **seven pre-specified quality categories**: correctness, completeness, conciseness, consistency, clarity, redundancy, and prominence/emphasis. Each category comprised multiple sub-criteria scored on a **0–3 scale** (0 = not present; 1 = poor; 2 = acceptable; 3 = fully compliant with regulatory expectations). For each category, **sub-criterion scores were summed and divided by the category-specific maximum** to yield a **normalized percentage score (0–100%)**, enabling comparison across categories with different numbers of sub-criteria. A **critical regulatory error** was predefined as any misrepresentation or omission likely to alter regulatory interpretation of safety, efficacy, or compliance (e.g., incorrect NOAEL attribution; omission of mandatory good laboratory practice (GLP) dose-formulation analysis).

**Results:** Use of the AutoIND capabilities in The Weave Platform reduced initial drafting time by 97% (from ~100 hours to 3.7 hours for 18,870 pages of 61 IND-1 source documents; ~100 hours to 2.6 hours for 11,425 pages of 58 IND-2 documents). Artificial Intelligence (AI)-generated content achieved quality scores of 69.58% and 77.85% for IND-1 and IND-2, respectively by normalizing the total scores to the maximum possible scores. No critical regulatory errors were detected. However, the experienced regulatory writing assessor identified deficiencies in the emphasis of the overall narrative which requires refinement from a human expert, either directly editing the content or instructing AI to update the content. Length (conciseness) and characteristic AI language (clarity) were also identified as potential improvements for a first draft.

**Conclusions:** LLM-based AI platforms such as AutoIND can dramatically accelerate preparation of the first drafts of regulatory documents such as written summaries of an IND. However, expert regulatory writing still is required to mature the output and the AI templates to submission-ready quality regarding emphasis and conciseness. The systematic nature of identified deficiencies provides a clear roadmap for improving AI performance in regulatory document generation.

# Introduction

The transition from **preclinical research to first-in-human (FIH)** clinical trials represents a critical inflection point in drug development. According to industry analysis, streamlining preclinical workflows can reduce the time to FIH by **40% or more**, accelerating drug entry into clinical trials—from approximately 24 months down to 12–15 months—and unlocking substantial value and patient access potential (McKinsey & Company, 2023).

In parallel, broader drug development automation is beginning to pay dividends. Experts predict that AI-driven approaches to drug discovery and safety assessment could **halve development timelines and costs** in the next three to five years (Reuters, 2025). Such reductions open the prospect of bringing more therapies to patients and extending commercial exclusivity windows.

Despite these gains upstream, one area remains unchanged: the **best-in-class preparation of IND applications**, especially written summaries of nonclinical findings. These regulatory documents remain heavily manual and time intensive. In many cases, they must be compiled rapidly once data arrive—representing a critical bottleneck in advancing candidate evaluation into FIH clinical trials (IQVIA Institute for Human Data Science, 2023).

LLMs offer one compelling solution. Their demonstrated ability to **generate clinical summaries** up to **28 times faster** than human writers—with comparable accuracy—suggests they could significantly streamline IND drafting (Biesheuvel et al., 2024; World Economic Forum, 2024; Quanticate, 2024).

To our knowledge, the use of LLMs to **accelerate IND document drafting**, and thereby FIH progression, has not yet been systematically evaluated. Accordingly, this study quantifies both time efficiency and quality outcomes of using an LLM-based tool—AutoIND—in producing written nonclinical summaries compared with traditional benchmarks. We hypothesize that AI-assisted workflows can reduce first-draft preparation time by **>70%**, potentially enabling faster IND filings and earlier clinical study starts for small and mid-sized biotech developers.

## INDs – Transforming Lab Breakthroughs into Clinical Reality

Preparing an Investigational New Drug (IND) application to support first-in-human (FIH) testing is a large-scale effort that requires compiling thousands of pages of preclinical study reports, manufacturing information, and clinical protocols (U.S. Food & Drug Administration, 2015). For modalities such as small and large molecules INDs often include 50–70 individual study reports; reflecting the breadth of nonclinical, pharmacology, pharmacokinetic, and toxicology data required. Recent reviews of regulatory digitalization note that drafting and verifying a single clinical trial application can consume hundreds of hours and months of coordinated cross-functional work (Ahluwalia et al., 2025). Timelines are further challenged by late-arriving experimental results—such as ongoing toxicology or pharmacokinetic studies—that must be integrated rapidly into already complex dossiers. These factors make IND preparation one of the most time-intensive activities in early drug development, underscoring the need for innovations that improve efficiency while preserving rigor and compliance.

## Promise and Limits of Large Language Models (LLMs)

Advances in LLMs have prompted renewed interest in automating regulatory writing. LLMs can summarize complex biomedical information and draft coherent text at unprecedented speeds. For example, recent studies show that LLM-generated clinical note summaries can be non-inferior to physician-written versions in completeness and correctness while being produced up to 28 times faster (Biesheuvel et al., 2024). Such findings have encouraged the development of domain-specific tools; for instance, the AutoIND tools in the Weave Platform use LLMs to generate draft IND sections from structured and unstructured source documents to cut initial drafting time drastically. These tools offer the potential to relieve some of the manual burden on regulatory writers and help teams manage compressed timelines.

## Risks and Governance

Despite their promise, integrating generative AI into regulated workflows is not without risks. LLMs can "hallucinate," producing plausible but incorrect or unverifiable information (Huang et al., 2024; Biesheuvel et al., 2024). Beyond outright errors, deficiencies in **correctness** (misinterpretation of results or numerical inaccuracies), **completeness** (omission of key methods or results), or **emphasis** (focusing on the wrong data while underreporting critical findings) can mislead reviewers and erode confidence. Likewise, poor **clarity** or non-scientific writing, excessive **redundancy**, and lack of **conciseness** can bog down reviewers, lengthen evaluation timelines, and delay clinical development.

These risks are compounded by concerns around data integrity, confidentiality, traceability, and alignment with regulatory guidance. Recognizing these challenges, the FDA's January 2025 draft guidance on the use of artificial intelligence in drug and biologics regulation sets out a **risk-based credibility assessment framework**, reflecting regulators' awareness of AI's potential while underscoring the need for robust oversight and systematic quality controls (U.S. Food & Drug Administration, 2025).

# Current Study

Despite increasing enthusiasm for AI in regulatory writing, systematic, real-world evaluations investigating the value to biotechnology companies remain scarce. To address this gap, Takeda Pharmaceuticals and Weave conducted a benchmarking study leveraging AI LLM-model AutoIND , *a template within the Weave Platform* assessing the preparation of written nonclinical summaries (eCTD Modules 2.6.2, 2.6.4, 2.6.6) for two historical INDs (IND-1: a large molecule with 61 discrete reports, June 2021; IND-2: a large molecule with 58 discrete reports, January 2024). Both cleared by the Food and Drug Administration within the last 4 years. Experienced regulatory writers (with at least 6 years' experience) used standardized rubrics to assess preparation time and document quality. The goal was to evaluate how well and how quickly LLM-based tools can generate a first draft for human-in-the-loop acceleration of written summary content.

The assessment identified the LLM-model significantly reduced time to first draft preparation but required subsequent expert regulatory writing to mature the content to submission-ready quality.

# Methods

## Study Design

This study evaluated AI-generated regulatory content using two historical INDs from Takeda (**IND-1, submitted June 2021; IND-2, submitted January 2024**). Source documents consisted of pharmacology, pharmacokinetics, and toxicology study reports organized under **Module 4 (M4) of the electronic Common Technical Document (eCTD) (**see *International Council for Harmonisation (ICH)*, https://www.ich.org/page/ctd)

For **IND-1**, 61 individual study reports were included, comprising approximately **18,870 pages**. For **IND-2**, 58 study reports were included, comprising approximately **11,425 pages**. Individual study reports typically ranged from **50 to several hundred pages**, reflecting the variability in scope and level of detail required for different pharmacology, pharmacokinetic, and toxicology assessments.

Prior to formal evaluation, a **preparation phase** was conducted. An experienced regulatory writer and an AI engineer used **five representative study reports from each IND** to iteratively adjust and refine the AutoIND prompts. These reports were selected only for calibration purposes and were **excluded from all subsequent quality assessments** to avoid bias. This ensured that the final evaluation reflected true performance of the AutoIND system on unseen source documents.

Drafts were generated using the **Weave AutoIND platform (version 2024-11-01)**. Drafting times for AutoIND were **directly recorded across all generation tasks**, providing measured efficiency metrics. PDF document upload and content extraction processing times are included in the recorded time. For context, **manual drafting times were estimated** based on the experience of regulatory writers (≥ 6 years' experience) at Takeda. These historical INDs, written manually, had been **reviewed and cleared by FDA** within the prior four years. By contrast, the AutoIND-generated drafts used for this analysis were **not submitted to the FDA**.

The AI-generated summaries were then assessed by a regulatory medical writer (≥ 6 years' experience) using a **comprehensive quality assessment framework**. (Figure 1)

The assessment framework focused on seven key quality dimensions, each containing multiple higher-resolution sub-criteria: **Correctness** (accuracy of scientific information and data representation relative to source documents), **Completeness** (thoroughness of content coverage per regulatory requirements), **Consistency** (internal coherence and alignment with established writing conventions), **Redundancy** (absence of unnecessary repetition or duplicated content), **Conciseness** (efficient communication without excessive verbosity), **Clarity** (readability and comprehensibility for regulatory audiences), and **Prominence/Emphasis** (appropriate highlighting of critical safety and efficacy information).

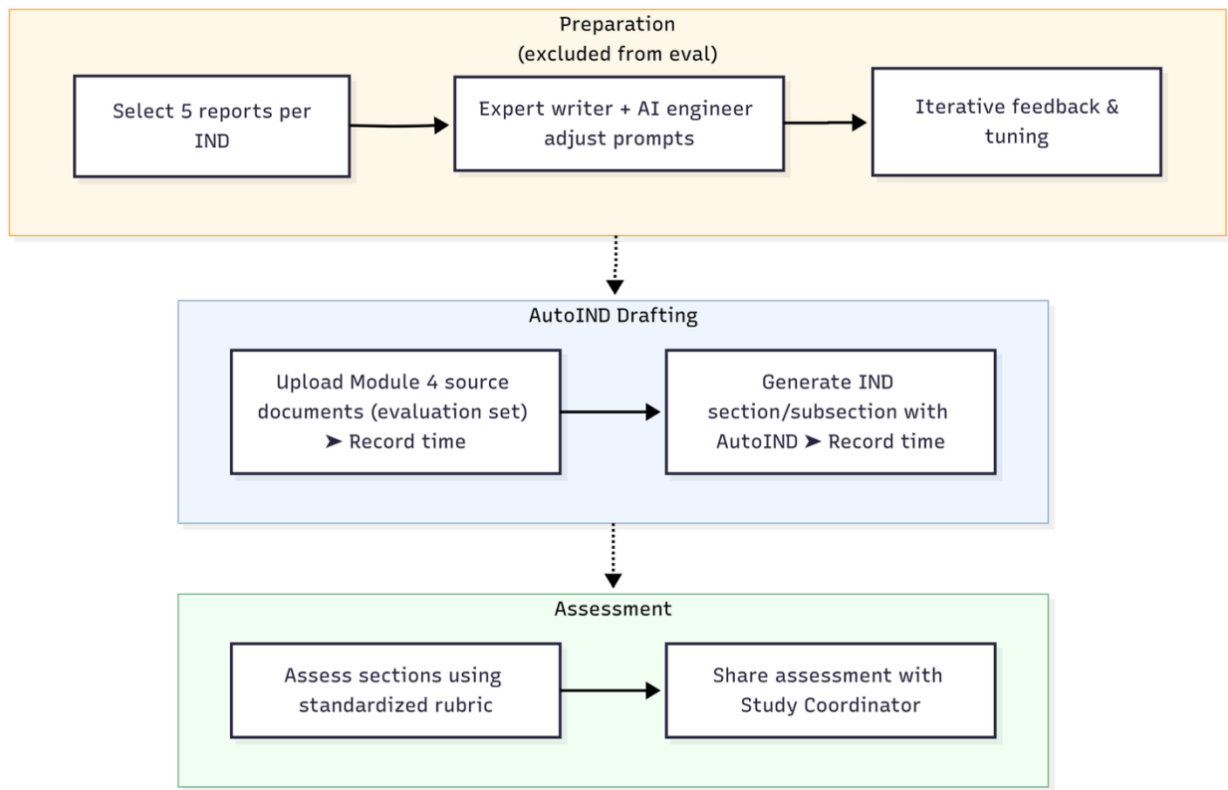

**Fig 1:** Process diagram for collaboration of medical writer and AutoIND.

Each criterion was evaluated using a standardized scale **0–3 scale** (0 = not present; 1 = poor; 2 = acceptable; 3 = fully compliant with regulatory expectations) with detailed scoring rubrics to ensure consistent assessment across all document sections (see Table 1 and Table S1). For each category, **sub-criterion scores were summed and divided by the category-specific maximum** to yield a **normalized percentage score (0–100%)**, enabling comparison across categories with different numbers of sub-criteria. A **critical regulatory error** was predefined as any misrepresentation or omission likely to alter regulatory interpretation of safety, efficacy, or compliance (e.g., incorrect NOAEL attribution; omission of mandatory GLP dose-formulation analysis).

Following quantitative scoring, an expert panel conducted detailed qualitative analysis of the AI-generated content, identifying systematic patterns and error archetypes across each quality dimension (see Supplementary Materials S2-S8 for comprehensive analyses).

## AI System Configuration

Weave Platform's AutoIND V2.3 (gpt-4-turbo, release 2024-10) operated inside a VPC on AWS. Source document PDFs were extracted using AWS Textract, classified into the IND sections, and the content was inserted into prompts customized to match Takeda's style guide.

## Documents & Tasks

Writers produced nonclinical written summaries (eCTD, Modules 2.6.2, 2.6.4, 2.6.6). IND 1 comprised 61 source files (18,870 pages); IND 2 comprised 58 files (11,425 pages).

## Outcome Measures

Primary: drafting time (minutes) recorded by Toggl. Secondary: quality score (0-3) for each subcategory of correctness, completeness, consistency, redundancy, conciseness, clarity, emphasis recorded by the regulatory assessor (Supplementary Table S1).

# Results

The study evaluated AI-assisted IND writing for nonclinical written summaries (eCTD, Modules 2.6.2, 2.6.4, 2.6.6) across two primary dimensions: quantitative metrics (time efficiency and quality scores, see Table 1, Table S1) and qualitative expert analysis of systematic patterns. While the quantitative results demonstrate impressive efficiency gains, the qualitative analysis reveals important nuances requiring consideration for practical implementation.

# Time Efficiency

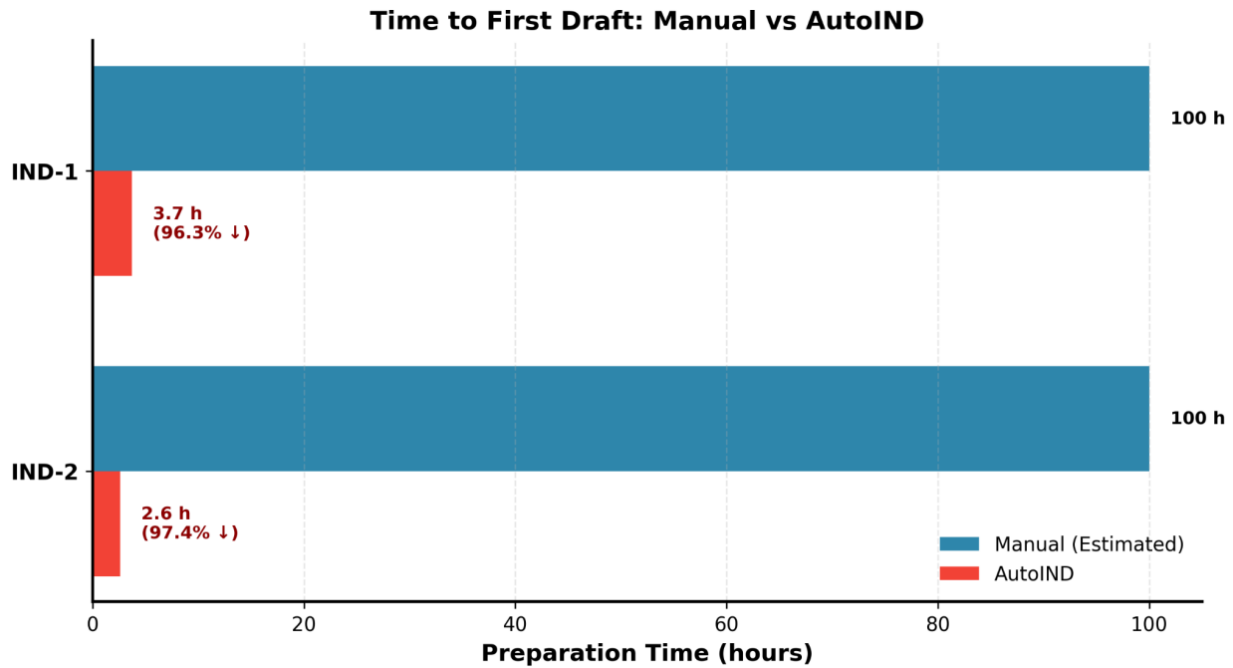

**Fig 2.** Comparison of time required for written summary generation between traditional manual writing and AI-assisted drafting using AutoIND. Traditional estimates based on historical writing time for comparable IND submissions. AutoIND times represent actual measured drafting time for initial complete drafts. Error bars represent standard deviation. Statistical significance: d = 6.8, p < 0.001.

The primary endpoint of drafting time reduction was dramatically achieved across both test INDs for written summary generation. Weave Platform's AutoIND demonstrated remarkable efficiency in producing initial written summary drafts, reducing preparation time by approximately 97% compared to traditional manual writing approaches (Figure 2).

**IND-1 Results:** AutoIND tools generated complete first drafts of written summaries in 3.7 hours from 61 source documents (18,870 pages), compared to an estimated 100 hours required for manual preparation. This represents a 96.3% time reduction.

**IND-2 Results:** AutoIND tools produced first drafts of written summaries in 2.6 hours from 58 source documents (11,425 pages), versus approximately 100 hours for traditional writing. This constitutes a 97.4% time reduction.

For written summary generation, mean pages-per-hour increased from 0.2 to 12.1.

These findings substantially exceed the hypothesized 70% time reduction, demonstrating that AI-assisted drafting can deliver exceptional efficiency gains in regulatory written summary preparation.

# Quality Assessment

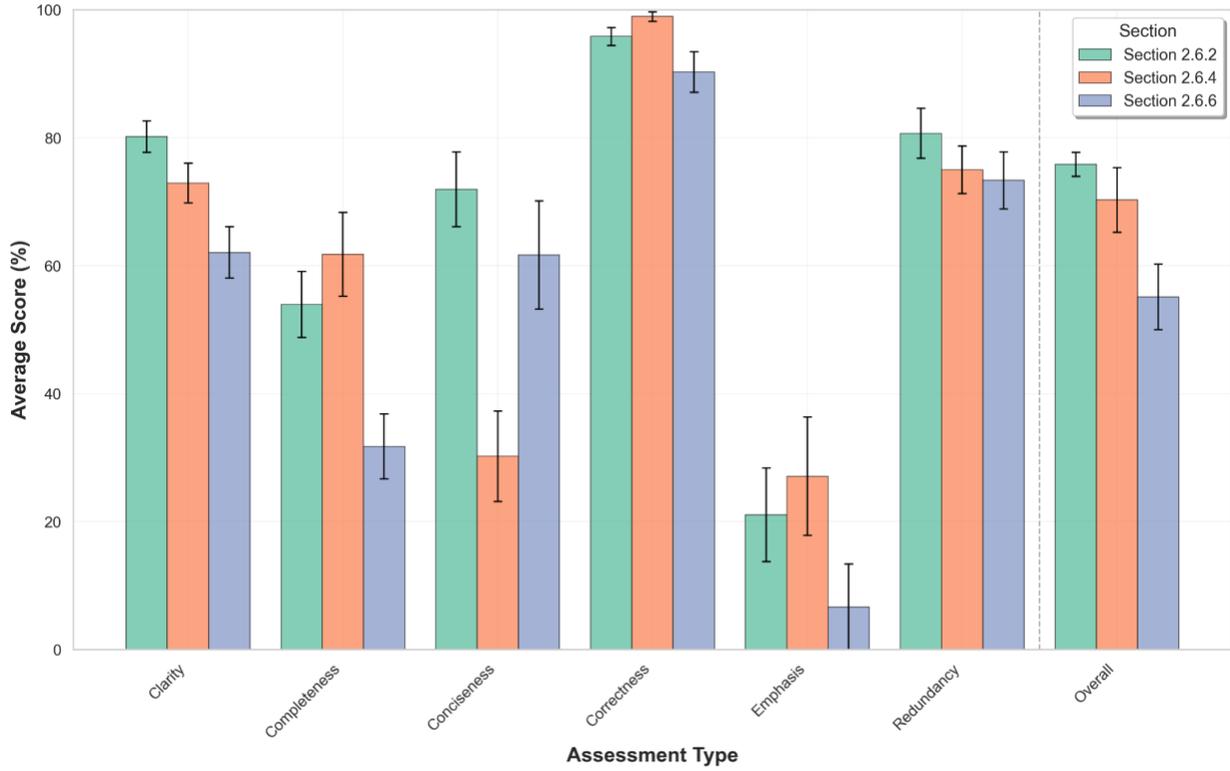
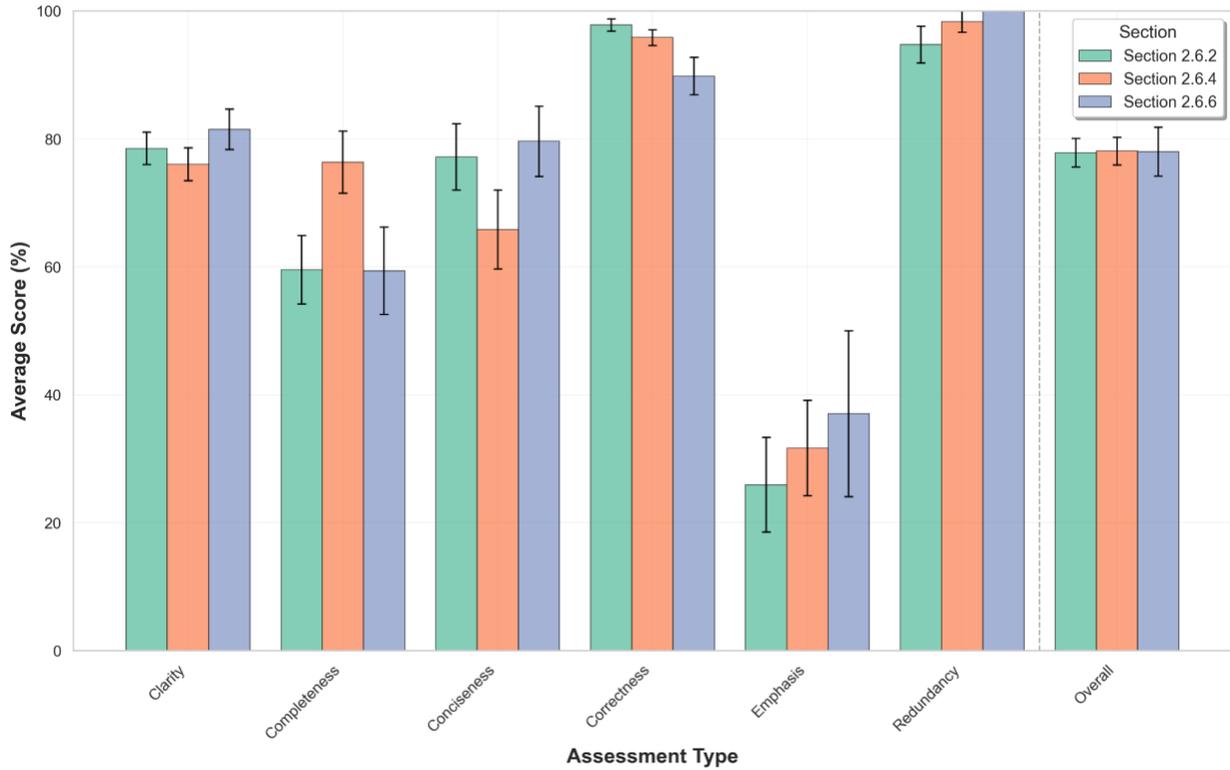

***Figure-3****:* Average percentage scores across criteria and module 2 section for IND-1 (a), and IND-2 (b) INDs. Error bar indicates the standard deviation. Y-axis represents the normalized score as percentage, calculated by dividing the total score for a criterion by the total possible score. The percentage score is not relative to human level.

When the results for **pharmacology (2.6.2)**, **pharmacokinetics (2.6.4)**, and **toxicology (2.6.6)** IND written summaries are examined separately, a few patterns stand out:

- **Correctness** was consistently high (>89% across all modules and INDs), suggesting that AutoIND largely preserved factual accuracy from the source documents regardless of content type.
- **Clarity** and **Redundancy** also remained relatively strong and consistent across modules, reinforcing that structural readability and avoidance of duplication are less variable between sections.
- **Emphasis** scored uniformly low, especially in toxicology (6.7% for IND-1, 37% for IND-2). This indicates that AutoIND struggled to highlight the most critical findings, often over-describing methods and under-weighting results — a systematic weakness.
- **Completeness** and **Conciseness**, however, varied more substantially. Pharmacokinetics sections (2.6.4) tended to have higher completeness (61% for IND-1, 76% for IND-2) but sometimes poor conciseness (30% for IND-1, 66% for IND-2). Toxicology (2.6.6) fared worst in completeness (32% for IND-1; 59% for IND-2), which is expected given the complexity and length of toxicology reports.

This pattern suggests that quality was broadly **similar across modules in terms of correctness, clarity, and redundancy**, but **systematically weaker in emphasis and variably weaker in completeness and conciseness**. That variability likely reflects differences in the structure and density of source reports: toxicology studies are more detailed, method-heavy, and variable in design, whereas pharmacology and pharmacokinetics are somewhat more standardized.

The AutoIND-assisted draft achieved an overall quality score of 69.58% and 77.85%, for IND-1 and IND-2, respectively. In other words, 10-25 % of generated content required revision or refinement.

Further quantitative deficiency analysis of the documents demonstrated the following:

- **Conciseness:** 3-5x word count inflation
- **Completeness:** 35% of summaries missing essential study design elements, 100% omission rate for GLP-required dose formulation analysis
- **Redundancy:** Backbone information, such as drug mechanism, analytical methods, etc., are repeated multiple times per document
- **Clarity:** 32% structural/organizational issues, 28% AI-characteristic language patterns

No critical regulatory errors were detected; however, detailed expert review (Supplementary Materials S2-S8) revealed systematic quality issues:

# Error Analysis and recommended Human Refinement Patterns

Detailed expert analysis of AI-generated content across seven quality dimensions revealed systematic patterns requiring human intervention (see Supplementary Materials S2-S8 for comprehensive analyses):

**Correctness (Supplementary Material S2)**

While **no critical regulatory errors** were detected, analysis identified several important patterns:

- **Mis-interpretation of results**: For example, AI claimed dose-response relationships where none existed and confused pharmacological parameters (e.g., "latency vs duration")
- **Hallucinated regulatory conclusions**: AI inappropriately assigned NOAEL (No Observed Adverse Effect Level) values to in-vitro studies and described adverse effects where none were reported
- **Numerical inaccuracies**: Wrong value (11 h vs 7 h onset, "PP increased …" vs "PP decreased")

**Completeness (Supplementary Material S3)**

Expert reviewers identified systematic omissions rendering documents unsuitable for regulatory submission:

- **Underspecified study design:** Missing N/group, vehicle composition, and dose regimens
- **Missing critical results data**: Absent toxicokinetic data, control group results, and quantitative values ($IC_{50}$/$EC_{50}$)
- **Minimal methods sections:** Complete omission or critical gaps in analytical methods and procedures

**Consistency (Supplementary Material S4)**

Analysis revealed significant standardization failures:

- **Uneven section detail**: Brief summaries too thin while conclusions verbose
- **Order/format drift**: Species ordering and dose presentation varied randomly
- **Terminology misalignment**: Multiple aliases for same compound (e.g., "GIPR agonist" vs "GIP receptor analogue")

**Redundancy (Supplementary Material S5)**

Systematic repetition patterns degraded document quality:

- **Endpoint echo-chamber**: Identical endpoint lists repeated verbatim in consecutive paragraphs
- **Method mantra**: Buffer compositions and analytical methods restated multiple times
- **Dose & route redundancy**: Administration details repeated throughout documents

**Conciseness (Supplementary Material S6)**

Quantitative analysis revealed severe verbosity:

- **3-5x word count inflation** compared to expert-written documents

- **Methods novella pattern**: Excessive procedural detail including buffer recipes and equipment specifications

**Clarity (Supplementary Material S7)**

156 clarity-related comments revealed:

- **Structural/organizational issues** (32%): Illogical information flow with results appearing before methods
- **Language/style problems** (28%): AI-characteristic phrases like "The study aimed to evaluate..." (12 occurrences)
- **Technical deficiencies** (15%): Unit formatting errors and inconsistent terminology

**Prominence/Emphasis (Supplementary Material S8)**

Critical information was systematically de-emphasized:

- **Method-heavy/result-thin** (>60% mentions): Long experimental details overshadowed key findings
- **Missing control data**: Making dose-response claims unverifiable
- **Generic qualifiers**: "At higher doses" used instead of specific numerical data

Analysis of the AI-generated content revealed specific patterns in the types of refinements typically required:

- **Structural reorganization:** 60-70% of the cases required repositioning for logical flow
- **Content reduction:** 70-80% of the cases needed reduction to achieve appropriate conciseness
- **Terminology standardization:** Consistent correction of regulatory terminology misuse

The expert panel noted that while no single error would necessarily compromise regulatory acceptance, the cumulative effect of these systematic deficiencies would require expert intervention.

Table 1 outlines the recommended human refinement patterns identified throughout the study.

# Discussion

## Primary Findings and Industry Impact

This study demonstrates that AI-assisted regulatory writing can dramatically accelerate IND preparation while maintaining acceptable quality standards. The 97% reduction in initial drafting time of the written summaries validates the transformative potential of generative AI in drug development. These efficiency gains dwarf prior LLM studies and validate the hypothesis that AI can handle the "70% cognitive lift" of regulatory writing—the systematic compilation and organization of technical data—freeing human experts for higher-value strategic interpretation and complex scientific judgment.

# AI Performance Patterns: Capabilities and Systematic Limitations

Our detailed expert analysis reveals a paradoxical performance pattern. While AI maintained consistent attention to detail and achieved remarkable speed, it exhibited systematic deficiencies across multiple quality dimensions that create a distinct "AI signature."

**Verbosity Paradox**: AI generated 3-5x longer documents than human-written equivalents, with systematic repetition patterns including "endpoint echo-chambers" and characteristic phrases like "The study aimed to evaluate..." appearing repeatedly. This verbosity masks critical information and wastes reviewer time.

**Completeness Gaps**: Despite lengthy output, AI systematically omitted critical regulatory requirements, with 100% failure rate for specific elements like dose formulation analysis in GLP studies and inappropriate assignment of NOAEL values to in-vitro studies.

**Structural Issues**: AI struggled with document architecture, exhibiting 32% failure rate in maintaining logical information flow and tendency to place results before methods, indicating fundamental misunderstanding of regulatory document structure.

These systematic patterns create immediately identifiable AI-generated content that undermines credibility with regulatory reviewers. However, all these issues can be addressed by providing additional refinement instruction to the AI, which is the expected workflow in such Regulatory Automation Management Platforms (RAMPs), such as Weave Platform. Furthermore, the consistent nature of these limitations presents clear remediation opportunities through specialized fine-tuning, structured generation templates, and hybrid approaches combining rule-based validators with generative capabilities.

| Quality Category | Common Issues | Recommended Human Refinement Patterns |
|---|---|---|
| **Correctness** | Misinterpretation of results, numerical inaccuracies | Detailed numeric verification, context clarification |
| **Completeness** | Missing methods details, omitted critical results | Adding explicit study design elements, numerical data |
| **Conciseness** | Excessive repetition, redundant methods | Removal of repeated boilerplate and unnecessary details |

| Quality Category | Common Issues | Recommended Human Refinement Patterns |
|---|---|---|
| **Consistency** | Variability in terminology, formatting | Harmonizing terminology, standardizing formatting |
| **Clarity** | Structural illogic, verbosity | Reordering sections, simplifying sentence structures |
| **Redundancy** | Repeated lists and statements | Eliminating repeated content, merging lists |
| **Emphasis** | Over-detailed methods, underreported outcomes | Reducing methods content, highlighting critical findings |

**Table-1:** Identified common issues and how human refinement can mitigate them

## Strategic Advantages Over Alternative Solutions

The findings strongly support AI adoption for accelerating first-draft composition, not because of labor cost reductions, but because of the **dramatic time efficiencies** achieved. AutoIND reduced drafting time by approximately **97%**, a level of acceleration that cannot be matched through conventional scaling of human writers. Importantly, by limiting the number of writers needed to mature AI-generated drafts, the final documents can maintain a **more consistent narrative voice and style**. In contrast, INDs written by large teams of medical writers often reveal variability in tone, terminology, and emphasis across sections. AI-assisted drafting allows human experts to concentrate on higher-order refinements—clarifying emphasis, ensuring completeness, and correcting terminology—while preserving **uniformity across the full dossier**. This combination of speed and consistency positions AI-enabled workflows as strategically superior to traditional approaches.

## Implications for Workforce Development and Regulatory Evolution

Such capabilities, like AutoIND, demonstrate potential value as a training tool for junior medical writers by providing structured examples of regulatory writing standards. For established companies, AI-assisted writing allows senior writers to focus on interpretive tasks while ensuring consistent quality across team members regardless of experience level.

The regulatory landscape is moving toward AI-enabled review processes. The U.S. Food and Drug Administration's January 2025 draft guidance on AI in regulatory decision-making introduces a risk-based framework for evaluating AI models, signaling the agency's openness to data-driven approaches (US Food and Drug Administration, 2025). In parallel, government initiatives outside of pharma illustrate the broader shift to data-first submissions. For example, the FedRAMP "20x" overhaul proposes automating validation for roughly 80% of cloud-security requirements, reducing the narrative portion of submissions to about 20% (Bourne and Alvarado, 2025). Life-sciences regulators and industry groups have likewise been advocating for more structured, data-centric submissions to improve efficiency and data integrity. AI-driven authoring tools are well positioned to support this transition, provided they address current challenges around conciseness, completeness, and factual reliability—issues that remain active areas of research and regulatory scrutiny.

## Innovation Opportunities and Future Directions

The systematic nature of identified limitations suggests clear paths for improvement:

1. specialized fine-tuning on regulatory documents with explicit verbosity penalties,
2. structured generation with enforced templates and mandatory field validation, and
3. domain-specific evaluation metrics beyond general language quality scores.

The success of AI-assisted IND writing indicates broad applicability across regulatory document types, including global marketing authorization applications, regulatory responses, safety reports, and protocol amendments. The core workflow demonstrated here—document ingestion, content generation with source fidelity, quality assessment, and compliance checking—can be adapted industry-wide for significant efficiency improvements beyond IND applications.

# Limitations

## Study Design Limitations

This evaluation focused on evaluating the preparation time and quality of the first draft of AutoIND document generation. In real-world applications, both AI and human-generated content undergo multiple revision cycles. A more comprehensive assessment would compare final polished documents from both approaches, though the dramatic time savings demonstrated here suggest AI would maintain its advantage even after multiple iterations.

The study was limited to two IND applications from a single therapeutic area, modality and pharmaceutical company (Takeda), potentially limiting generalizability across different therapeutic areas, modalities, company styles, and document types. Future studies should include broader representation across the pharmaceutical industry to confirm these findings.

Our expert panel analysis (Supplementary Materials S2-S8) revealed systematic AI deficiencies that may not fully manifest in quantitative quality scores. For instance, while AI achieved 69.58% and 77.85% quality scores relative to the expected submission-ready final version quality, the detailed review identified pervasive issues with verbosity (3-5x inflation), completeness (35% missing critical information), and clarity (32% structural issues) that would require substantial human intervention.

## Evaluation Limitations

The use of a single assessor, while consistent, introduces potential bias in quality evaluation. Future studies should employ multiple independent evaluators to ensure robust quality assessment. Additionally, the subjective nature of some quality criteria (particularly clarity and emphasis) may limit the precision of comparative scoring.

The study did not assess long-term regulatory outcomes, such as FDA review comments or approval timelines, which would provide ultimate validation of document quality. Follow-up studies should track regulatory success rates for AI-assisted submissions.

## Technology Limitations

This evaluation was specific to the AutoIND tools in The Weave Platform and may not generalize to other AI writing tools. The rapid evolution of AI technology suggests that capabilities and performance will continue to improve, potentially making these findings conservative estimates of AI potential.

The study's focus on nonclinical written summaries (eCTD Modules 2.6.2, 2.6.4, 2.6.6) may not reflect performance across other IND sections, such as nonclinical tabulated summaries, nonclinical overall summary, clinical protocols or manufacturing information, which have different complexity and standardization requirements.

# Conclusion & Outlook

This study provides robust evidence that AI-assisted regulatory writing can address critical pharmaceutical industry challenges around efficiency and capacity. The dramatic 97% reduction in initial drafting time demonstrates transformative potential, though our comprehensive expert analysis (Supplementary Materials S2-S8) reveals significant quality challenges that must be addressed.

The findings support a human-AI collaboration model where artificial intelligence handles systematic compilation and synthesis tasks while human experts provide essential oversight. Our analysis identified systematic deficiencies including:

- 3-5x word count inflation requiring aggressive editing
- 35% of documents missing essential study design elements

Despite these challenges, the consistency of deficiency patterns suggests targeted improvements could significantly enhance AI performance. The expert panel's identification of specific error archetypes (e.g., "endpoint echo-chambers," "method mantras," "impressionist summaries") provides a clear roadmap for system refinement.

AI-augmented writers produce IND drafts in hours, but achieving regulatory-ready quality requires addressing the identified deficiencies. We propose a five-step scalable framework:

1. **Ingest/Extract** secure documents and data
2. **Draft** section-specific content via LLM in minutes
3. **Refine/ Review** content with subject matter expert judgments and systematic error correction
4. **Verify** by tracing information flow to sources and checking completeness
5. **Publish/Monitor** eCTD build and AI-drafted responses

Beyond immediate efficiency gains, AI-assisted writing positions pharmaceutical companies for success in an evolving regulatory environment where agencies increasingly employ AI for review processes. The standardization and machine-readability inherent in AI-generated content will become increasingly valuable as regulatory frameworks continue to modernize.

The broader implications extend beyond IND applications to the entire spectrum of regulatory documentation. As AI capabilities continue advancing, the pharmaceutical industry can fundamentally transform how it approaches regulatory compliance, potentially accelerating drug development timelines while improving document quality and consistency.

These findings contribute essential empirical evidence for the responsible adoption of AI in highly regulated industries. While current AI systems demonstrate impressive efficiency gains, our expert analysis reveals they generate **surface-level drafts** rather than submission-ready documents. Achieving regulatory-grade AI will require advances across multiple dimensions:

1. **Immediate technical fixes:** Word count limits, mandatory field validation, unit standardization.
2. **Structural improvements:** Template enforcement, cross-reference validation, completeness scoring.
3. **Training enhancements:** Domain-specific fine-tuning on expert-annotated documents with penalties for verbosity and omissions.
4. **Quality control systems:** Automated detection of AI signatures and systematic deficiency patterns.
5. **Source document standardization:** Consistent data presentation, lean report structures, and adoption of templated study formats that minimize variability and redundancy, ensuring the AI has well-organized inputs to summarize accurately.

Future research should explore the scalability of these findings across different modalities, broader therapeutic areas and regulatory processes, while continuing to monitor the evolution of AI capabilities and regulatory frameworks. The evidence presented here supports cautious but confident adoption of AI-assisted writing tools as a strategic advantage in pharmaceutical development, provided organizations invest in the necessary refinements to address identified deficiencies.

Investment in validation frameworks and up-skilling is urged as regulators adopt AI review. The systematic nature of current AI deficiencies, while challenging, also provides a clear roadmap for creating next-generation systems capable of producing truly submission-ready regulatory documents.

# Ethics & Compliance

Data remained encrypted in transit and at rest; AutoIND parameters were not updated with sponsor data. The study adhered to WHO 2023 ethical AI principles and GAMP 5 guidelines.

# Materials & Data Availability

Redacted Takeda IND excerpts are available upon request.

# Author Contributions

- Conceptualization: UE, YG, AC, BR, AR
- Methodology: UE, YG, MG, LJS, AC, BR, SW
- Expert IND analysis: LJS
- Content formatting: LJS, SW
- Investigation: UE, BR, SW
- Writing – Original Draft: UE
- Writing – Review & Editing: All authors
- Funding Acquisition: Takeda & Weave

# Funding

Supported jointly by Takeda Pharmaceutical Co. Ltd. and Weave, Inc. No external funding.

# Conflicts of Interest

Weave authors are employees and shareholders of Weave; Takeda authors are shareholders of Takeda. LJS is a consultant and not a shareholder of either company. The study did not involve commercial bias in design or analysis.

# Acknowledgments

We thank the Takeda Nonclinical Regulatory Systems and Writing (NRWS) Department, especially Marian Glynn, for timeline data and scorecard development, Tammy Lyons for reviewing the manuscript.

# Figures & Tables

- Figure 1: Study workflow diagram
- Figure 2: Draft-time waterfall chart
- Figure 3: Bar plot of quality aspects
- Table 1: Quality aspects and examples

*Note: Detailed expert analyses of quality dimensions with specific examples and quantitative breakdowns are provided in Supplementary Materials S2-S8.*

# Supplementary Material

- Table S1: Full QC rubric
- Supplementary Material S2: Expert Analysis - Correctness Evaluation
- Supplementary Material S3: Expert Analysis - Completeness Evaluation
- Supplementary Material S4: Expert Analysis - Consistency Evaluation
- Supplementary Material S5: Expert Analysis - Redundancy Evaluation
- Supplementary Material S6: Expert Analysis - Conciseness Evaluation
- Supplementary Material S7: Expert Analysis - Clarity Evaluation

- Supplementary Material S8: Expert Analysis - Prominence/Emphasis Evaluation
- Prompt templates for AutoIND generation
- Raw timeline logs